\documentclass[letterpaper, 10 pt, conference]{ieeeconf}  

\IEEEoverridecommandlockouts                              

\usepackage{graphicx} 
\usepackage{float} 
\usepackage{subfig}
\usepackage{algorithm}
\usepackage{array}
\usepackage{algorithmic}
\usepackage{amsfonts}
\usepackage{amsmath,amssymb}
\usepackage{mathrsfs}
\usepackage{cite}
\usepackage{balance}
\usepackage{bbding}
\usepackage{stfloats}
\usepackage{float}

\usepackage{booktabs}
\usepackage{multirow}
\usepackage{amssymb}
\usepackage{pifont}
\usepackage{makecell}
\usepackage[table]{xcolor} 
\usepackage{xcolor}
\newcommand{\cmark}{\ding{51}}%
\newcommand{\xmark}{\ding{55}}%
\title{\LARGE \bf
Enhancing Vision-Based Policies with Omni-View and Cross-Modality Knowledge Distillation for Mobile Robots 
}

\author{ 
     Kai Li$^{1,2}$, Shiyu Zhao$^{2}$ 
     \thanks{ This research work was supported by National Natural Science Foundation of China (Grant No. 62473320). (Corresponding author:Shiyu Zhao.)}
	\thanks{$^{1}$College of Computer Science and Technology at Zhejiang University, Hangzhou, China.}
	\thanks{$^{2}$School of Engineering at Westlake University, Hangzhou, China.
	{\tt\small \{likai,zhaoshiyu\}@westlake.edu.cn}}%
    \thanks{$^{3}$ Code is at this link: https://github.com/xiaowei1015/robot-kd  }
}

\usepackage[mathscr]{eucal}
\begin{document}

	\maketitle
	\thispagestyle{empty}
	\pagestyle{empty}

	\begin{abstract}
Vision-based policies are widely applied in robotics for tasks such as manipulation and locomotion. On lightweight mobile robots, however, they face a trilemma of limited scene transferability, restricted onboard computation resources, and sensor hardware cost. To address these issues, we propose a knowledge distillation approach that transfers knowledge from an information-rich, appearance-invariant omni-view depth policy to a lightweight monocular policy. The key idea is to train the student not only to mimic the expert’s actions but also to align with the latent embeddings of the omni-view depth teacher. Experiments demonstrate that omni-view and depth inputs improve the scene transfer and navigation performance, and that the proposed distillation method enhances the performance of a single-view monocular policy, compared with policies solely imitating actions. Real-world experiments further validate the effectiveness and practicality of our approach. Code will be released publicly$^{1}$.

	\end{abstract}

	\section{Introduction}
Vision-based policies for mobile robots have a wide range of applications, from underwater exploration\cite{manderson2020vision}, to quadruped pursuit-evasion\cite{bajcsy2024learning} and drone flight\cite{wu2024whole,geles2024demonstrating}. By directly mapping raw image data to control actions, these policies eliminate the need for explicit intermediate representations such as maps and trajectories\cite{wu2024whole}, and offer a more streamlined approach to vision-based robotic tasks.
Among the various methods for training visuomotor policies, imitation learning (IL) has become a popular paradigm. Its high sample efficiency and straightforward formulation make it well-suited for training visuomotor policies from expert demonstrations\cite{zhuang2024enhancing, cheng2024extreme,zhuang2023robot,manderson2020vision,bajcsy2024learning}.

However, for mobile robots, visuomotor policies face two major challenges.
\textbf{First}, visuomotor policies with monocular cameras often struggle with poor generalization to unseen environments. Due to limited data diversity in both simulation and real-world settings, the trained policy may fail when exposed to new visual observations that are unseen in the training set, leading to unpredictable actions and behaviors. Even within the same environment, variations in illumination and the texture of surrounding objects can negatively affect the performance of the policy.
\begin{figure}[thpb]
      \centering
      \includegraphics[scale=0.29]{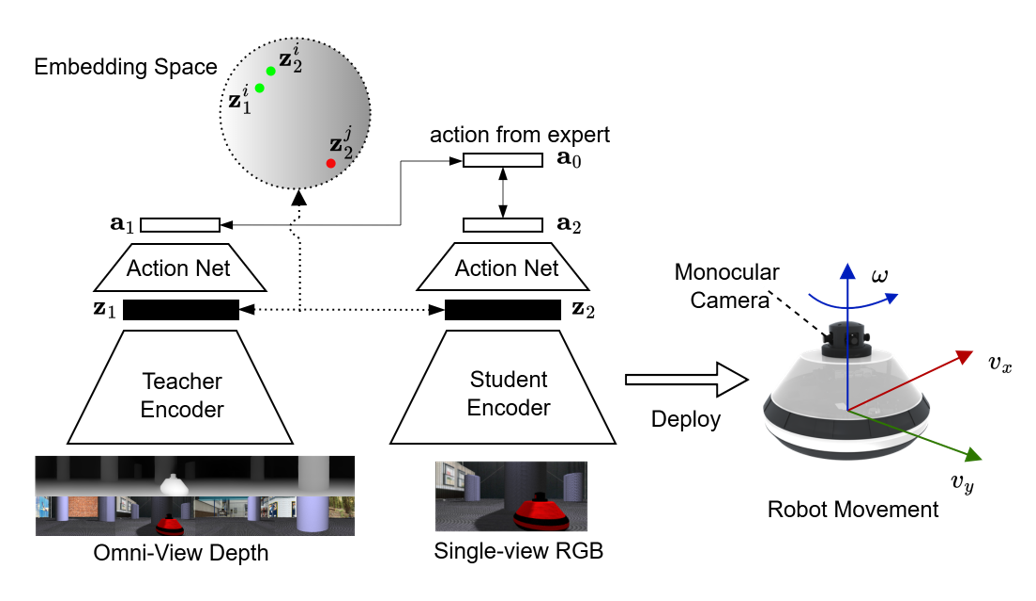}
      \caption{ The teacher policy (left) leverages appearance-invariant omnidirectional depth images generated by concatenating multi-view inputs, while the student (right) relies on a single-view RGB image. Both the teacher and the student imitate the action $\mathbf{a}$ from the expert data, with the student additionally distilling the feature embedding $\mathbf{z}$ from the teacher. Compared to the teacher, the student policy is computationally lightweight and more suitable for deployment on lightweight low-cost mobile robots.
      }
      \label{pipeline}
   \end{figure}
\textbf{Second}, mobile robots are typically equipped with limited onboard computing resources, due to constraints in power consumption and mechanical structure. 
Visuomotor policies for mobile robots often benefit from large field‑of‑view (FOV) image inputs, since these deliver a large volume of visual information about the surroundings. Omnidirectional robots particularly benefit from 360° perception for improved obstacle awareness and navigation safety\cite{wang2025omni,xu2022omni}. However, constrained onboard computational resources pose challenges for real-time processing of the increased data volume of omnidirectional sensing. In addition, high-capacity image encoders are often preferred in visuomotor policies for their ability to provide high-quality visual representations, but they also require significant onboard computational resources.

Using depth sensors, such as RGB-D cameras\cite{cheng2024extreme,zhuang2023robot} or LiDAR\cite{wang2025omni,xu2025flying}, can provide appearance-invariant depth observations, which improves the scene transfer ability of the policy. While effective, the high price and bulky size of depth sensors limits their application in low-cost, lightweight mobile robots. Alternatively, high-capacity monocular depth networks\cite{depth_anything_v2} can produce depth images without additional hardware and enhance the policy's scene transfer ability, but their computational demands exacerbate the existing challenge of limited onboard computation resources.
   
The preceding discussion about the trilemma of 1) a policy's scene transfer and safety performance, 2) onboard resource limitation, and 3) robot hardware cost, naturally leads to the following question: \textbf{is it feasible to learn a policy from a low-cost single-view RGB camera that achieves performance close to that of policies using depth images and omnidirectional views?} Motivated by this question, we propose a knowledge distillation framework to enhance vision-based policies trained from imitation learning by distilling the omni-view and cross-modality information.
Contributions of this paper are summarized as follows:

1) We introduce an omni-view and cross-modality knowledge distillation framework for mobile robots utilizing vision-based IL. Compared to other methods that take monocular images, policies trained using our approach achieve approximately 15\% improvement in navigational success rate, an approximate 19\% increase in collision-free travel distance, and a reduction in action errors.

2) In online deployment, our method eliminates the need for depth sensors or multi-camera omnidirectional systems, with onboard inference taking around 20 ms. This not only reduces the computational load on the robot but also lowers hardware costs and complexity. 

3) Extensive experiments are conducted in both simulated and real-world environments. The deployment on a real-world lightweight mobile robot system with limited computation resources demonstrates the practical effectiveness and feasibility of our approach.

\section{Related Works}

\textbf{Visuomotor Policy Learning}.
Visuomotor policies directly map raw pixel observations to robot actions without the need for intermediate representations such as state estimation or trajectories\cite{geles2024demonstrating,wu2024whole}, which have emerged as a promising approach in various robotic tasks such as quadruped locomotion \cite{zhuang2023robot,cheng2024extreme} and drone flight \cite{geles2024demonstrating,wu2024whole,zhang2025learning}. 
Due to its high sample efficiency and straightforward formulation, imitation learning has become a popular paradigm for training visuomotor policies.
To enhance the performance and scene transferability of these policies, researchers have explored several methods. Some works leverage depth images\cite{cheng2024extreme,zhuang2023robot} or visual attention areas\cite{zhuang2024enhancing,liu2020using,liang2024visarl} to improve performance. The work in\cite{xing2024contrastive} leverages adaptive contrastive learning (ACL) to learn more discriminative visual features from monocular images. In the context of reinforcement learning (RL), the work in \cite{zhang2025learning} uses cross-modality information to improve monocular vision policy performance.

\textbf{Knowledge Distillation}.
Knowledge distillation \cite{hinton2015distilling} was originally proposed to transfer knowledge from a large model to a compact one. In robot policy learning, this corresponds to distilling a teacher policy into a student. While imitation learning typically trains the student to replicate the teacher’s actions, recent works also distill intermediate embeddings to improve student performance.
In quadruped locomotion, several studies \cite{bajcsy2024learning, kumar2021rma, lee2020learning} distill knowledge from a fully observable teacher to a student with partial observations. The student’s encoder is trained to learn the teacher’s latent representations, which capture information inaccessible to the student, such as an evader’s predicted trajectory \cite{bajcsy2024learning} or terrain and mechanical parameters \cite{kumar2021rma, lee2020learning}. 
In robot manipulation\cite{acar2023visual, chen2024trakdis} or monocular depth estimation\cite{han2024boosting}, some works\cite{acar2023visual, chen2024trakdis,han2024boosting} apply a similar principle to transfer knowledge from a teacher with information-rich input to a student with less informative input.

In this paper, we distill the robust appearance-invariant and information-rich omnidirectional depth knowledge to a compact policy that takes only a single-view RGB image for mobile robot navigation. By combining the cross-modality and  omni-view distillation, we seek to improve the scene transfer and navigational success performance of the mobile robot, under the condition of limited onboard computation resources and robot hardware cost.

\begin{figure*}[thpb]
      \centering
      \includegraphics[scale=0.42]{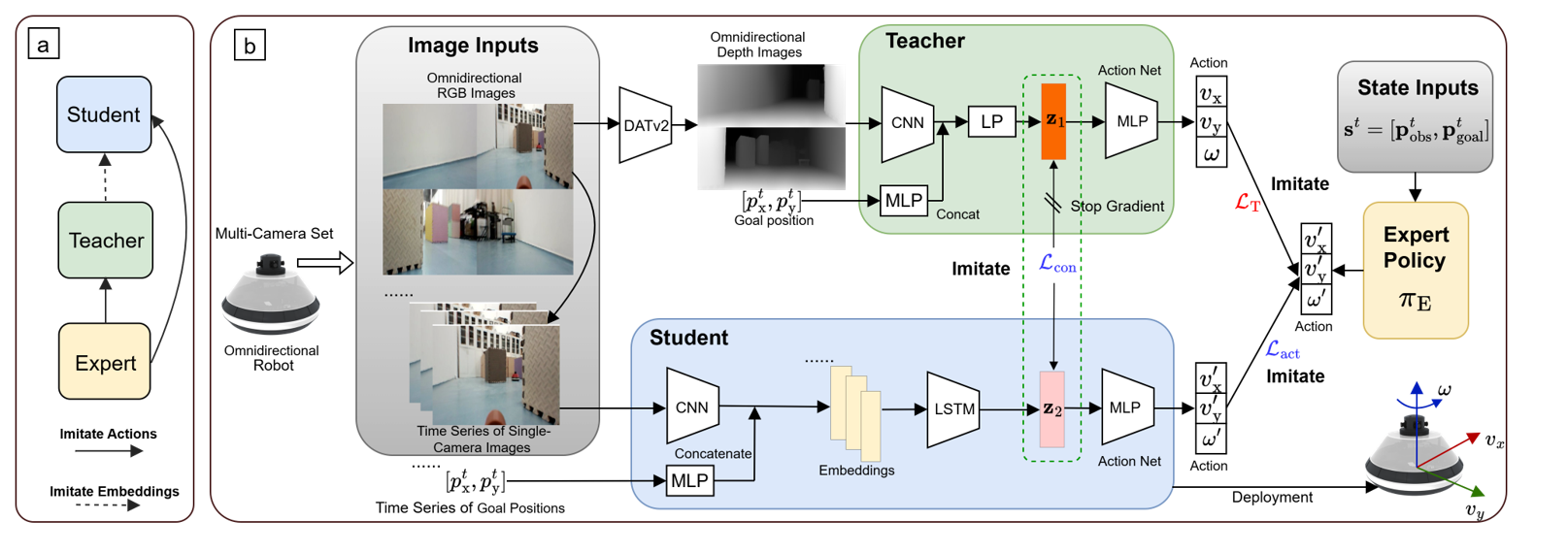}
      \caption{Overview of the proposed method. (a) shows the knowledge transfer flow from the state-based expert, to the omnidirectional-depth-based teacher, and the RGB-based student. (b) shows the detailed pipeline of our method. In contrast with the vanilla form of IL, our method imitates both the action output and intermediate visual embeddings, which is highlighted in the dash-line boxes. LP in the teacher denotes linear projection, which is used for embedding dimension matching.}
      \label{pipeline}
   \end{figure*}
   
\section{Proposed Method}
\subsection{Problem Formulation}
Our work aims to learn a visuomotor policy that receives a temporal sequence of image observations and goal states, and outputs control actions. The image sequence and goal states are encoded into a joint embedding $\mathbf{z}^t$, which the policy uses to produce control actions. The policy is formulated as:
\begin{align}
\mathbf{a}^t = \pi(\mathbf{z}^t),
\end{align}
where $\mathbf{z}^t$ denotes the latent embedding at time step $t$, and $\mathbf{a}^{t}=[v_{\rm{x}}^{t},v_{\rm{y}}^{t},\omega^{t}]$ is the 2D velocity command. The goal state $\mathbf{g}^{t}$ represents a 2D goal point in the robot's local coordinate frame.
A key challenge with this approach is that using a monocular camera with a limited FOV can constrain the policy's scene transferability and collision avoidance capabilities, especially when trained with limited data. To address these limitations, we introduce our knowledge distillation framework in the following sections.

\subsection{Overview}
Fig.~\ref{pipeline} provides an overview of our method. We first use an off-the-shelf expert policy $\pi_{\text{E}}$ to collect demonstration data. Based on this data, we train a teacher policy $\pi_{\text{T}}$ that uses omnidirectional depth images as raw input. The latent embeddings of $\pi_{\text{T}}$ and actions of $\pi_{\text{E}}$ are then distilled into a student policy $\pi_{\text{S}}$, which operates with single-view RGB input. Finally, $\pi_{\text{S}}$ is deployed onboard to perform the task. In the following sections, we describe the architectures and training of the vision-based policies $\pi_{\text{T}}$ and $\pi_{\text{S}}$, the contrastive embedding distillation, and how we obtain the expert policy $\pi_{\text{E}}$.
\subsection{Teacher Policy Training}
We first employ an off-the-shelf expert policy $\pi_{\text{E}}$ to collect expert demonstrations in the form of (state, observation, action) tuples, where the state comprises the robot’s global position $[x, y]$ and heading $\psi$, and the observation is an omnidirectional image obtained by concatenating images from the onboard multi-camera setup. 
Using expert data, we train the teacher policy. Since our lightweight, low-cost mobile robot is equipped with only monocular cameras, we first convert omnidirectional RGB images to depth images using DepthAnyThingV2 (DATv2) \cite{depth_anything_v2}. The omnidirectional depth images are then sent to an encoder network, which produces a depth image embedding. This depth embedding is concatenated with an embedding of the goal point, and linearly projected to a fused embedding $\mathbf{z}_{1}$. Finally, $\mathbf{z}_{1}$ is processed by a multilayer perceptron (MLP) network to generate the final action. $\pi_{\text{T}}$ is trained with a vanilla imitation loss, which minimizes the $L_{2}$ discrepancy between its output action and the expert's action,
\begin{equation}
\mathcal{L}_{\text{T}} = \| \pi_{\text{T},\theta_{\text{T}}}(\mathbf{z}_{1}^t) - \pi_{\text{E}}(\mathbf{s}^t)\|_{2}.
\label{L0}
\end{equation}
Here, the policy inputs $\mathbf{z}_{1}^{t}$ refer to the linearly projected embedding from concatenated embeddings of omnidirectional image and goal point. $\theta_{\text{T}}$ denotes the learnable parameters of $\pi_{\text{T}}$. $\mathbf{s}^{t}$ denotes the state inputs for $\pi_{\text{E}}$. Since there are no available image encoders designed for omnidirectional images with a large aspect ratio, we divide the omnidirectional detph image into four separate parts evenly. Each part is then processed by a shared image encoder, and the resulting embeddings are concatenated to form a final, joint embedding. The image encoder, goal state encoder and the action network are trained end-to-end with no frozen modules. 

\subsection{Student Policy Training}
The student takes a time series of RGB images from a single-view monocular camera. This image sequence is processed by a shared image encoder to generate a series of image embeddings. Simultaneously, the goal point state is encoded by an MLP. At each time step, the image embedding is concatenated with the goal point embedding to form a series of fused embeddings. These fused embeddings are then passed to a long short-term memory (LSTM) recurrent module for temporal fusion. The resulting embedding $\mathbf{z}_{2}$, which contains both spatial and temporal information, is then fed into a final MLP to regress the action output. This approach leverages historical data to overcome the limited information of a single static image, providing the policy with a richer understanding of the robot's surroundings.
The student is trained using the following loss function,
\begin{align}
\mathcal{L}_{\text{S}} = \lambda_{0}\mathcal{L}_{\text{act}} + \lambda_{1}\mathcal{L}_{\text{con}}
\label{eq_Ls}
\end{align}
where $\mathcal{L}_{\text{act}}$ is the loss for action regression and $\mathcal{L}_{\text{con}}$ is the contrastive loss for aligning the feature embeddings. The hyperparameter $\lambda_{0}$ and $\lambda_{1}$ balance the two loss components. The action loss, $\mathcal{L}_{\text{act}}$, minimizes the $L_{2}$ discrepancy between the action outputs of $\pi_{\text{S}}$ and $\pi_{\text{E}}$,
\begin{align}
\mathcal{L}_{\text{act}} = \| \pi_{\text{S},\theta_{\text{S}}}(\mathbf{z}_{2}^{t})- \pi_{\text{E}}(\mathbf{s}^{t}) \|_{2}.
\label{eq:L1_act}
\end{align}
The inherent limitations of a single-view RGB input, including its limited information volume and sensitivity to appearance changes, make it difficult for the student to generalize. As a result, action imitation (Eq.~\eqref{eq:L1_act}) alone fails to ensure robust policy performance.
Therefore, we introduce an additional contrastive loss $\mathcal{L}_{\text{con}}$ to align the student's embeddings with those of the teacher. 
The contrastive loss and feature embedding alignment will be introduced in detail in Section III-E. During training of the student, the image and state encoders, LSTM, and action network are trained end-to-end without any frozen modules.

\subsection{Knowledge Distillation via Contrastive Learning} 
The key part of our knowledge distillation framework is the alignment of intermediate features in the embedding space. Advances in visual representation learning and knowledge distillation\cite{chen2020simple,tian2019contrastive} have demonstrated that contrastive learning loss is effective in pulling together feature embeddings of similar samples while pushing apart those of dissimilar ones. This mechanism enables the learning of high-quality discriminative representations that better capture the underlying structure of the data. Based on these considerations, we use the InfoNCE contrastive loss \cite{chen2020simple} to align the teacher's embedding $\mathbf{z}_{1}$, with the student's embeddings $\mathbf{z}_{2}$. 
The loss function is defined as
\begin{equation}
\mathcal{L}_{\text{con}} = \frac{1}{N} \sum_{i=1}^{N}
-\log
\frac{
\exp\left(\mathrm{sim}(\mathbf{z}_{1}^i, \mathbf{z}_{2}^i)/\tau\right)}{
\sum_{j=1}^{N} \exp\left(\mathrm{sim}(\mathbf{z}_{1}^i, \mathbf{z}_{2}^j)/\tau\right)}.
\end{equation}
Here, $\tau$ is a temperature parameter that shapes the similarity distribution, while $\mathrm{sim}$ is a similarity measurement for the embeddings. $N$ represents the batch size. In this formulation, positive pairs are embeddings corresponding to robot states that are spatially close in the global pose, denoted as $(\mathbf{z}_{1}^i, \mathbf{z}_{2}^i)$. Conversely, negative pairs are sampled from embeddings associated with spatially distant states, 
and represented as $(\mathbf{z}_{1}^i, \mathbf{z}_{2}^j)$.
We use cosine similarity, a measurement commonly used in contrastive learning because of its scale invariance and stability, to measure embedding similarity,
\begin{equation}
\mathrm{sim}(\mathbf{z}_{1}, \mathbf{z}_{2}) = \frac{ \mathbf{z}_{1}^T \mathbf{z}_{2} }{ \| \mathbf{z}_{1} \| \| \mathbf{z}_{2} \| }.
\end{equation}
The InfoNCE loss pushes the embedding $\mathbf{z}_{2}$ of the student toward the embedding $\mathbf{z}_{1}$ of the teacher when sampled at spatially close states, and pushes them apart when sampled from distant poses.
Since $\mathbf{z}_{1}$ is produced from omnidirectional view depth images, it encodes rich appearance-invariant information about the surrounding environment. 
By using the contrastive loss to align $\mathbf{z}_{1}$ and $\mathbf{z}_{2}$, the student's embedding is enriched with two key types of information from the teacher: 1) the surrounding environment context not visible from a single-view camera, and 2) the appearance-invariant depth information.
The LSTM recurrent module in the student helps to accumulate temporal information for $\mathbf{z}_{2}$, which narrows the semantic gap between $\mathbf{z}_{1}$ and $\mathbf{z}_{2}$ and helps an easier and more stable embedding alignment.

\begin{figure}[thpb]
    \centering   
    \includegraphics[scale=0.44]{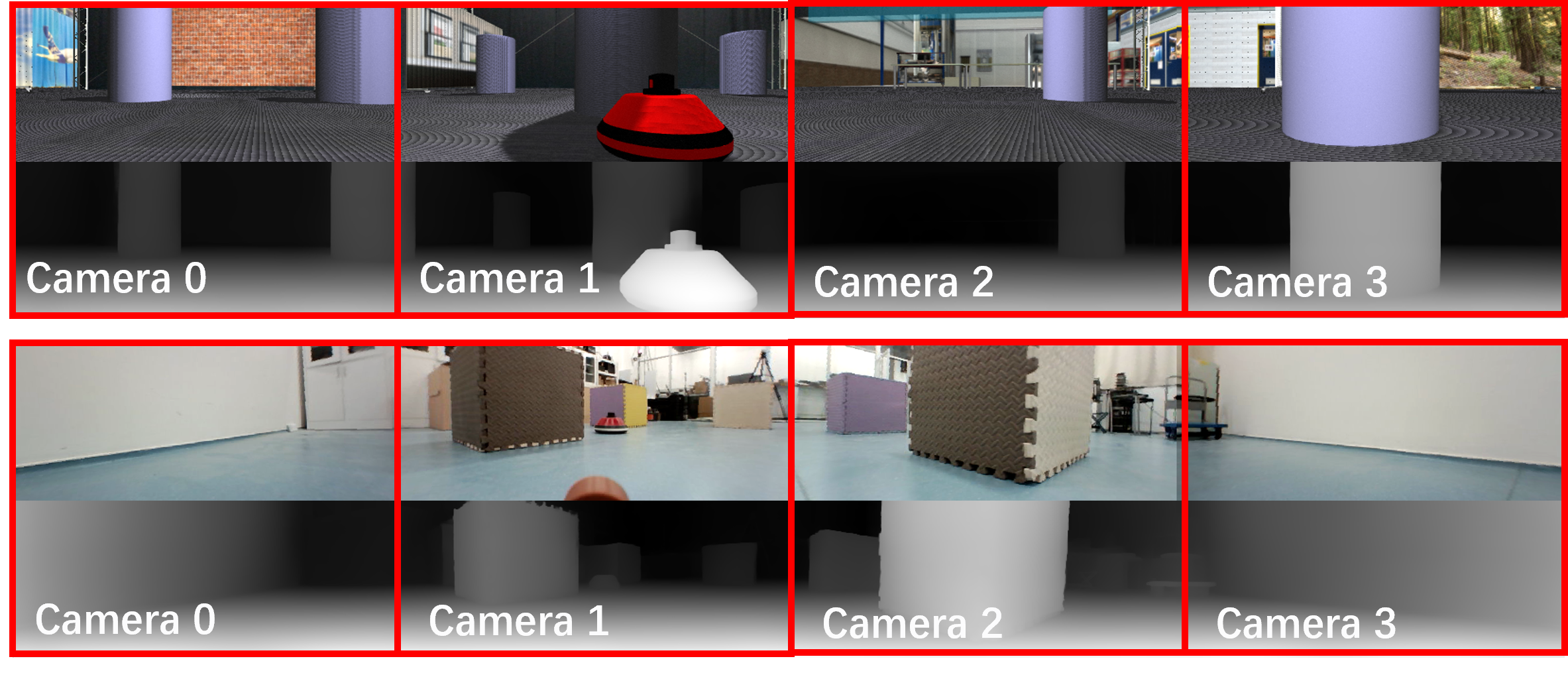}  
    \caption{
  Omnidirectional RGB and depth images from DATv2. The top row shows the simulation and the bottom row shows the real world. Omni-view is formed by concatenating multi-camera images. Camera 1 is used for the single-view student policy.
    }
    \label{figure_pursue}
\end{figure}

The omni-view and cross-modality knowledge distillation can be intuitively understood through two perspectives: \textbf{regularization} and \textbf{data augmentation}.
From a regularization standpoint, the contrastive loss $\mathcal{L}_{\rm{con}}$ acts as a regularizer for the imitation learning process. By adding it to the standard action loss $\mathcal{L}_{\rm{act}}$, the student is not only encouraged to mimic the action of the expert but is also constrained to learn a similar, robust intermediate representation from the teacher. This prevents the student from simply overfitting to the observation-action data pairs and forces it to learn a more generalizable feature space.
Alternatively, this can be viewed as a form of data augmentation. The single-view image, which is the student's input, can be considered a “cropped” or “augmented” version of the teacher's information-rich and appearance-invariant omnidirectional depth input. Our goal is to ensure that the policy achieves performance on this “augmented” input comparable to that on the original data.
By using contrastive learning to align the embeddings, the policy is effectively trained on the “augmented” data, which inherently improves its representation power and performance.
 
\subsection{Expert Policy for Data Generation}
In this section we introduce how the expert policy $\pi_{\rm{E}}$ is obtained. It is worth noting that $\pi_{\rm{E}}$ is not limited to a particular form, for example it can be via model predictive control (MPC)\cite{luis2019trajectory}, RL\cite{bajcsy2024learning,xing2024contrastive,wu2024whole}, or even human demonstrations\cite{manderson2020vision}. 
In this work, we choose to use state-based RL to learn $\pi_{\text{E}}$, since RL enjoys higher onboard inference speed than optimization-based methods like MPC.
$\pi_{\rm{E}}$ takes the states and outputs control commands. The state inputs are defined as $\mathbf{s}^{t}=[\mathbf{p}_{\rm{obs}}^t,\mathbf{p}_{\rm{goal}}^t] $, where $\mathbf{p}_{\rm{obs}}^t$ denotes
the relative positions of obstacles and $\mathbf{p}_{\rm{goal}}^t$ denotes the relative goal position in the local frame. The raw state vector is first encoded by a two-layer MLP and then sent into the actor and critic network. The reward at time $t$ is defined as $r^{t}=r_{\rm{dist}}^{t} + r_{\rm{obst}}^{t} + r_{\rm{bound}}^{t}$. Each component of the reward is defined as,
\begin{equation}  
\begin{split}    
r_{\rm{dist}}^{t} &= -\alpha\, \lvert d \rvert \\
r_{\rm{obst}}^{t} &= -10 \ \text{if hits the obstacles} \\  
r_{\rm{bound}}^{t} &= -10 \ \text{if hits the boundaries}.
\end{split}
\label{reward_componet_ally}
\end{equation}
where $\alpha$ is a hyperparameter, $d$ is the distance between the robot and goal point.

\section{Experimental Evaluations}
In this section, we first introduce the experiment task, environment and the implementation details, then we evaluate scene transfer performance with embedding analysis. Finally, we evaluate the performance in the robotic task.
\subsection{Task Description and Implementation Details}
\textbf{Task Description}. We evaluate our knowledge distillation method on the robot navigation task. The omnidirectional robot uses images and a goal point coordinate in its local coordinate frame as input, and navigates toward the goal point while avoiding collisions. The goal point may be either static or moving according to a predefined policy. The robot uses only its onboard camera for perception and decision-making. The robot's aim is to approach the goal point and maintain desired distance and heading angle.

\textbf{Data Description}. The training dataset is structured as a sequence of (state, image, action) tuples. Here the state is the global position and heading of the robot. Data are collected by rolling out the state-based expert policy $\pi_{\text{E}}$ in 4 different simulation environments and the real world lab environment. We build a simulation environment with ROS Gazebo\cite{koenig2004design}. The experiment area is 4 m $\times$ 4 m. The obstacles in the environment are randomly placed, and the obstacle density (obstacle area / total area) varies from 0.05 to 0.20.
The images are rendered with 4 onboard RGB cameras, each with 110$^\circ$ horizontal FOV and 30 Hz frame rate. We calibrate the 4 cameras, which are pointed in different directions. After calibrating each camera, we undistort the raw images based on the calibrated parameters and concatenate them to create a single omnidirectional view. 
The goal point information for the policy is provided by simulation or motion tracking system.
In total, we collected 50k (state, image, action) tuples for policy training.


   
\textbf{Policy Training}. 
The state-based expert policy $\pi_{\text{E}}$ is trained with the Soft Actor-Critic (SAC) algorithm \cite{haarnoja2018soft} for 5M environment steps. The omnidirectional depth teacher distills knowledge from $\pi_{\text{E}}$ and is trained with a batch size of 64. The learning rate is $1\times10^{-5}$ for the encoder and $1\times10^{-4}$ for the action network. Depth images are generated by DATv2 \cite{depth_anything_v2}, each single view having a resolution of $3\times360\times180$ pixels.
For the student policy, we use a batch size of 32, with the same learning rates as the teacher for the encoder ($1\times10^{-5}$) and the action network ($1\times10^{-4}$). The dimensions of $\mathbf{z}_{1}$ and $\mathbf{z}_{2}$ are both 1024.
The weighting factor of the contrastive embedding distillation loss in Eq.~\eqref{eq_Ls} ($\lambda_{1}$) is set to 0.9 at the beginning of training and linearly decayed to 0.1. This schedule allows the policy to initially focus on embedding alignment and gradually shift its emphasis toward action learning. The weighting factor $\lambda_{0}$ is fixed at 1.
The temperature parameter $\tau$ for contrastive learning is 0.1. The image encoder is ResNet \cite{he2016deep}, initialized from ImageNet-1K \cite{deng2009imagenet}. 
The student is trained with DAgger \cite{ross2011reduction} to improve generalization power. 

\subsection{Scene Transfer Performance}
\begin{figure}[thpb]
    \centering
    \includegraphics[scale=0.09]{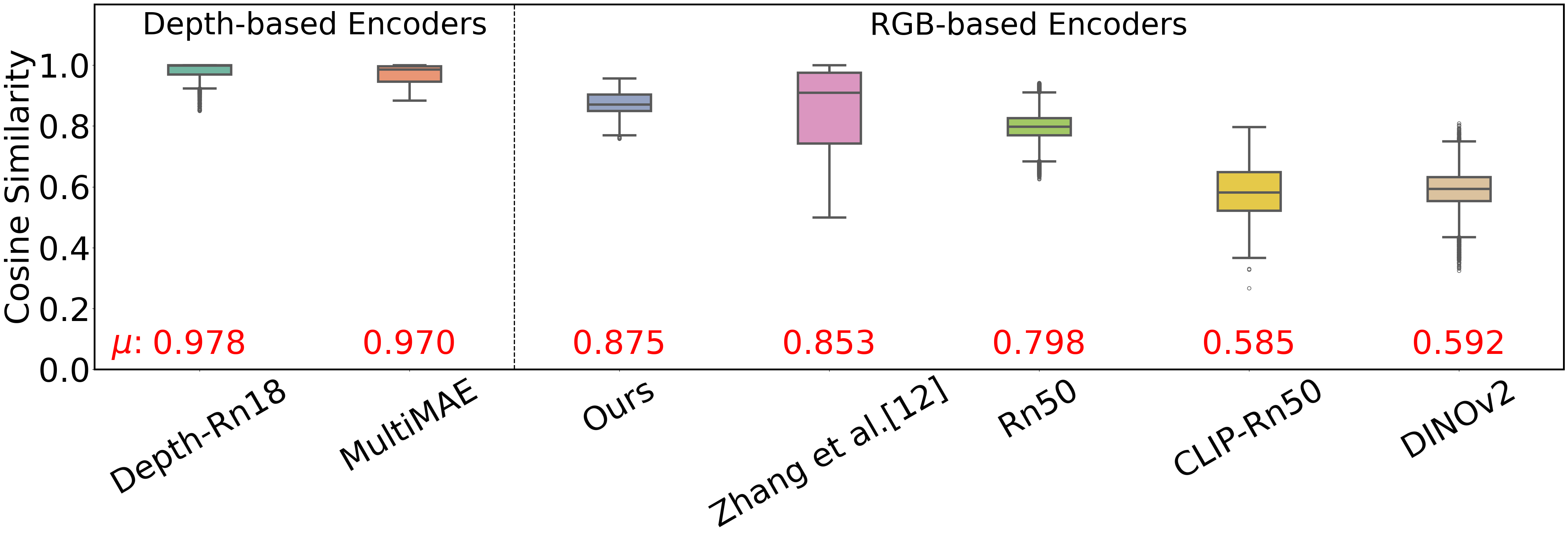}
    \caption{ Embedding similarity comparison of different image encoders across scenes. Higher similarity values indicate more consistent embeddings and stronger scene transferability. $\mu$ is the mean value of similarity. 
    }
    \label{figure_similarity}
\end{figure}
\begin{figure}[thpb]
    \centering
    \includegraphics[scale=0.28]{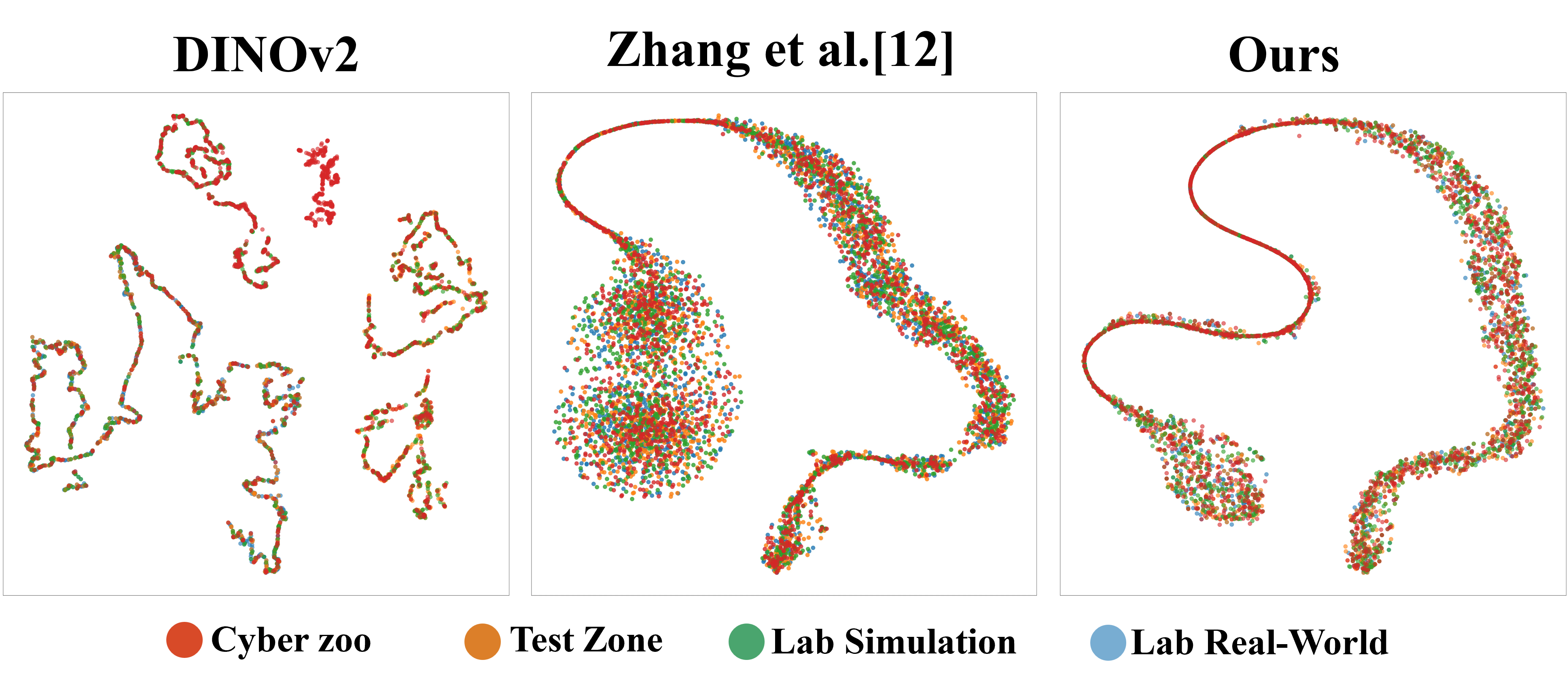}
    \caption{ t-SNE\cite{maaten2008visualizing} visualization of visual embeddings of different scenes from DINOv2\cite{oquab2024dinov2}, Zhang et al.\cite{zhang2025learning} and ours. Different colors represent different scenes.
    }
    \label{figure_tsne}
\end{figure}

\begin{figure*}[thpb]
      \centering
      \includegraphics[scale=0.75]{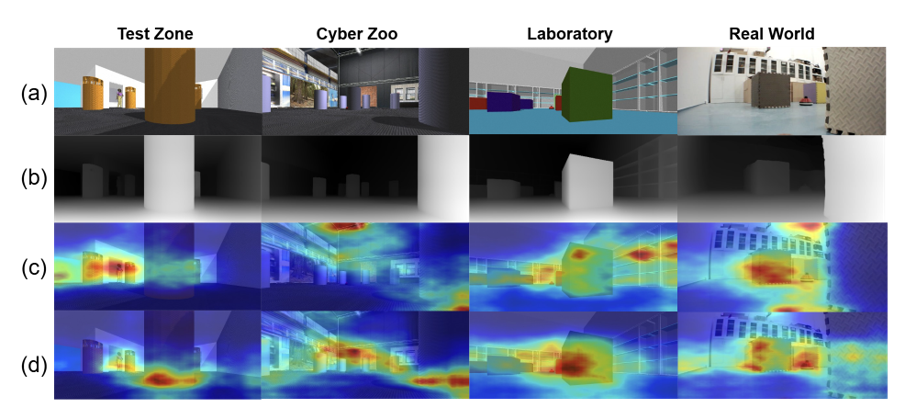}
      \caption{The rows show (a) RGB images, (b) depth images, (c) Full Gradient\cite{srinivas2019full} (FG) attention maps from a policy trained directly on RGB, and (d) FG attention maps from a policy trained with our knowledge distillation method. The attention maps in row (c) are distracted by irrelevant features. In contrast, the maps in row (d) align more with human intuition, consistently focusing on obstacles when they are nearby. The columns show different simulation and real-world scenes. The encoder backbones of (c) and (d) are both Resnet50. }
      \label{figure_FG}
 \end{figure*}

 \begin{table}[h]
\setlength\tabcolsep{4.7pt}
\caption{
White rows show action error results for methods without embedding distillation, using different input modalities and camera views. 
Gray rows show results with the proposed knowledge distillation method.
\textit{Finetuned} indicates that the encoder is further trained on our custom robot dataset, while non-finetuned encoders use frozen weights pretrained on a large-scale public dataset.
}
\begin{center}
\begin{tabular}{l c c c c c}
\toprule
\textbf{Method}    &  \textbf{Modality} & \textbf{View} & \textbf{Finetuned}   &   \makecell{ \textbf{AE} \\ \textbf{Train}}  $\downarrow$ &   \makecell{ \textbf{AE} \\ \textbf{Test}} 	$\downarrow$  \\
\midrule
Resnet18\cite{he2016deep}            & RGB      & Omni  & \xmark    & 0.0038  & 0.0226        \\
Resnet18\cite{he2016deep}            & RGB      & Single  & \xmark    & 0.0051  & 0.0734            \\
Resnet50\cite{he2016deep}            & RGB      & Omni  & \xmark    & 0.0027  & 0.0093      \\
Resnet50\cite{he2016deep}            & RGB      & Single  & \xmark    & 0.0042  & 0.0112      \\
DINOv2\cite{oquab2024dinov2}         & RGB      & Omni  & \xmark    & 0.0030  & 0.0097                      \\
DINOv2\cite{oquab2024dinov2}         & RGB      & Single  & \xmark    & 0.0049  & 0.0129                           \\
CLIP Rnet\cite{radford2021learning}  & RGB      & Omni  & \xmark    & 0.0031  & 0.0138                      \\
CLIP Rnet\cite{radford2021learning}  & RGB      & Single  & \xmark    & 0.0052  & 0.0143                        \\
MAE\cite{he2022masked}               & RGB      & Single  & \xmark    & 0.0036  & 0.0090                                         \\
Resnet18\cite{he2016deep}                           & Depth    & Omni  & \cmark    &   \textbf{0.0012}   & \textbf{0.0037}      \\
Resnet18\cite{he2016deep}                           & Depth    & Single  & \cmark    &  0.0020   & 0.0050                        \\
MultiMAE\cite{bachmann2022multimae}  & RGB-D    & Single  & \xmark    &  0.0029  & 0.0091                        \\\rowcolor{gray!20} 
\textbf{Ours-Rnet18}                        & RGB      & Single  & \cmark    &  0.0030    & 0.0070                                  \\\rowcolor{gray!20} 
\textbf{Ours-Rnet50}                        & RGB      & Single  & \cmark    &  0.0021    & 0.0067                                  \\

\bottomrule
\end{tabular}
\end{center}
\label{table_action_error}
\end{table}
\textbf{Visual Embedding Similarity}.
We evaluate the scene transfer ability of visuomotor policies by measuring the similarity of their visual embeddings across different scenes. To do this, we use distinct simulation environments that share the same layout, meaning the positions and sizes of obstacles, as well as the robot's start and goal positions, are identical. The only variable is the environment's visual appearance.
We use the average cosine similarity of the embeddings to quantify this transfer ability. Ideally, with identical scene layouts, test policies and initial robot states, the visual embeddings at the same position should be highly similar across different scenes.
As shown in Fig.~\ref{figure_similarity}, both the depth-image and RGB-D-based encoders (MultiMAE \cite{bachmann2022multimae}) achieve high embedding similarities. Among RGB-based encoders, our knowledge distillation framework attains the highest mean embedding similarity, indicating more consistent features across different scenes.
We also use t-SNE\cite{maaten2008visualizing} to visualize embeddings from different scenes. Fig.~\ref{figure_tsne} shows the visualization results. The embeddings of ours are more consistent across different scenes compared with other RGB-based encoders. Moreover, the embedding distribution aligns with the robot’s spatial movements and preserves meaningful spatial variation in the input images, ensuring that distinct observations are represented by distinct embeddings. In contrast, embeddings from pretrained models (e.g. DINOv2) scatter without clear structure. 

\textbf{Visual Attention Maps}. We use the Full Gradient method \cite{srinivas2019full} to generate visual attention maps. In Fig.~\ref{figure_FG}, we compare the attention maps from encoders trained with and without our knowledge distillation method.
The results demonstrate that our policy, trained with knowledge distillation, learns to focus its attention on task-relevant features, aligning more closely with human intuition. In contrast, the policy without knowledge distillation is easily distracted by irrelevant features in the RGB input, leading to a less robust representation.

\subsection{Mobile Robot Task Performance}
\textbf{Action Error.} We perform comparison and ablation studies using Action Error (AE) to quantify the discrepancy between expert and student outputs. Table~\ref{table_action_error} reports the AE of different visual encoders under various training conditions without embedding distillation (white rows), and with embedding distillation (gray rows).
For visual encoders that are not fine-tuned, we use pre-trained weights from large-scale public datasets. For a fair comparison, all policies incorporate temporal information using an LSTM with the same sequence length of 5 steps.
The results confirm that a larger volume of input information (i.e. omni-view input) and the depth modality are advantageous for policy performance. Specifically, the data show that using omnidirectional depth images produces the lowest AE. \textbf{When the input modalities are the same}, an omni-view input generally lowers AE in both training and testing scenes compared to single-view setups, which indicates improved policy performance. Similarly, \textbf{when the types of input view are the same}, using depth as the input modality leads to a lower AE, suggesting enhanced imitation performance.
The policy trained with our omni-view, cross-modality distillation achieves a lower AE than other RGB-based encoders, approaching the performance of omni-view depth input. This demonstrates the effectiveness of our framework in leveraging omni-view and cross-modality information to enhance visuomotor policies.

\begin{table*}[h]
\setlength\tabcolsep{4.2pt}
\caption{Results for the robot navigation task. All onboard models are accelerated with NVIDIA TensorRT using FP16 precision. For a fair comparison, all policies incorporate temporal information using an LSTM with the same sequence length of 5. The table reports Action Error (AE), Success Rate (SR), and Moving Distance (MD). The onboard time is presented as the mean, with the 99\% percentile shown in parentheses. The latency of the raw image is around 30 ms. For methods using depth as input, the inference time of DATv2 must be included in the total latency.}
\begin{center}
\begin{tabular}{l c c c c c c c c c}
\toprule
\textbf{Method}    &  \textbf{Modality}  & \textbf{View}  & \textbf{Finetuned}  &   \textbf{AE} $\downarrow$ &  \textbf{SR (\%)} $\uparrow$   & \textbf{MD (m)} $\downarrow$&  \textbf{\#Parameters} & \textbf{Encoder Time (ms)} $\downarrow$  & \textbf{DATv2 Time (ms)} \\
\midrule
Resnet18\cite{he2016deep}           & RGB      & Single   & \xmark   & 0.0739  & 31.0   &  4.08 ± 0.34   & 11.7M   &  16.51 (26.64)     & -                   \\
Resnet18\cite{he2016deep}           & RGB      & Omni   & \xmark   & 0.0479  & 40.0   &  5.19 ± 0.42   & 11.7M   &  21.36 (29.54)     & -                   \\
Resnet18\cite{he2016deep}           & Depth    & Single   & \cmark   & 0.0050  & 69.0   &  8.28 ± 0.33   & 11.7M   &  10.15  (17.62)     & 25.69 (34.88)       \\
Resnet18\cite{he2016deep}           & Depth    & Omni   & \cmark   & 0.0037  & 74.0   &  8.94 ± 0.31   & 11.7M   &  12.20 (19.90)     & 54.97 (82.20)       \\

Resnet50\cite{he2016deep}           & RGB      &  Single  & \xmark   & 0.0229  & 38.0   &  5.35 ± 0.21   & 23.5M   &  24.10 (33.81)     & -                  \\
Resnet50\cite{he2016deep}           & RGB      &  Omni  & \xmark   & 0.0115  & 49.0   &  7.09 ± 0.30   & 23.5M   &  45.34 (58.78)     & -                  \\
Resnet50\cite{he2016deep}           & Depth    &  Single  & \cmark   & 0.0069  & 70.0   &  8.87 ± 0.14   & 23.5M   &  15.52 (26.89)     & 25.69 (34.88)      \\
Resnet50\cite{he2016deep}           & Depth    &  Omni  & \cmark   & 0.0033    & 80.0   &  8.90 ± 0.28   & 23.5M   &  42.78 (60.30)     & 54.97 (82.20)     \\

CLIP Rnet\cite{radford2021learning} & RGB      &  Single  & \xmark   & 0.0165  & 40.0   &  5.13 ± 0.67   & 38.3M   &  24.98  (32.76)    & -                  \\
MultiMAE\cite{bachmann2022multimae} & RGB-D    &  Single  & \xmark   & 0.0130  & 49.0   &  5.92 ± 0.55   & 87.1M   &  43.97  (51.26)    & 25.97 (31.17)      \\
DINOv2\cite{oquab2024dinov2}        & RGB      &  Single  & \xmark   & 0.0158  & 34.0   &  4.85 ± 0.61   & 86.0M   &  132.45 (190.01)   & -                       \\
MAE\cite{he2022masked}              & RGB      &  Single  & \xmark   & 0.0095  & 50.0   &  6.25 ± 0.32   & 86.6M   &  35.68  (50.20)    & -                  \\ \rowcolor{gray!20}

\textbf{Ours-Rnet18}                & RGB      &  Single  & \cmark   & 0.0070  & 67.0   &  8.08 ± 0.21 & 11.7M     &  16.85 (26.61)     & -                  \\\rowcolor{gray!20} 
\textbf{Ours-Rnet50}                & RGB      &  Single  & \cmark   & 0.0067  & 72.0   &  8.49 ± 0.25 & 23.5M     &  24.10 (31.45)     & -                  \\

\bottomrule
\end{tabular}
\end{center}
\label{table_robot_task}
\end{table*}

\begin{figure}[thpb]
      \centering
      \includegraphics[scale=0.54]{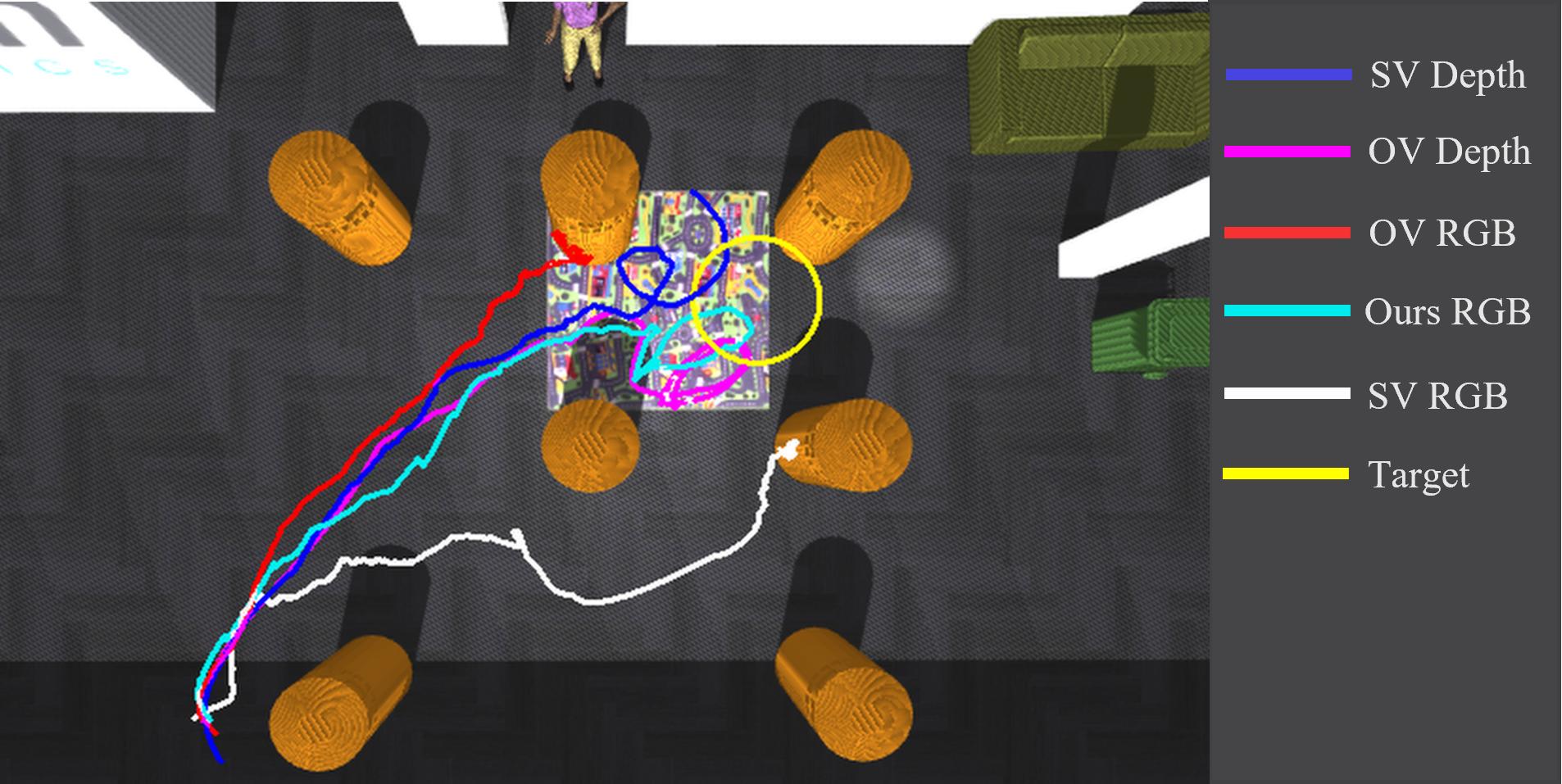}     
      \caption{Typical experiment trajectories of robots under policies trained with different strategies and input modalities. The target is moving in a circle (Yellow trajectory). SV denotes single-view image input, and OV denotes omni-view image input. }
      \label{figure_trajectory}
 \end{figure}

\textbf{Navigational Performance}. Table~\ref{table_robot_task} shows the results of the robot navigation experiment. AE is the action error in test scenes. The success rate (SR) is defined as the proportion of experiment episodes in which the robot reaches the goal within 0.3 m without collision. The moving distance (MD) is the distance the robot moves in one episode before collision with obstacles or boundaries.
We perform 80 independent runs in simulation (20 for each scene) and 20 runs in the real-world for each experiment.
The onboard computer is an NVIDIA Jetson Orin NX, featuring 6 ARM cores, 16 GB memory, 25 W power consumption, and up to 100 TFLOPS of computing performance. The onboard model is accelerated with TensorRT with FP16 precision. 
From the results in Table~\ref{table_robot_task}, we can see that using our knowledge distillation shows the best AE, SR and MD performance among RGB-based encoders, with a 23\% increase in SR and 20\% in MD. Since the robot is equipped with RGB cameras only, we use DATv2 to generate depth images onboard. 
Although using depth as input shows stronger performance in simulation, in the real-world, the additional latency (25.7 ms for single-view and 55.0 ms for omni-view) for depth image inference is undesirable for real-time control. Moreover, the CPU and GPU load introduced by a separate high-capacity depth estimation module further increases the onboard burden (with a 39\% GPU load increase and 19\% CPU increase).
High-capacity models such as DINOv2 takes longer inference time (more than 130 ms), which is not suitable for onboard running. In contrast, with our distillation method, the policy performance gets enhanced without relying on depth estimation module, which is particularly beneficial for mobile robots with limited onboard computation resources. Fig.~\ref{figure_trajectory} shows typical robot trajectories under different input modalities and views. Policies using raw RGB images collide with obstacles due to poor scene transfer. In contrast, depth input or RGB with our distillation method enables safe navigation and target tracking. Single-view depth input (blue trajectory) can still cause minor collisions when obstacles leave the field of view. 
 \begin{table}[h]
\setlength\tabcolsep{4.3pt}
\caption{Comparison with other methods in the literature that focus on enhancing the performance of visuomotor policies.}
\begin{center}
\begin{tabular}{l  c c c c }
\toprule
\textbf{Method}   &   \textbf{AE (Train)}$\downarrow$ &   \textbf{AE (Test)}$\downarrow$  &   \textbf{SR \%}$\uparrow$  & \textbf{MD (m)}$\uparrow$  \\
\midrule
RoboSaGA\cite{zhuang2024enhancing}     & 0.0088  &  0.0301  & 47.0 & 5.59 ± 0.51           \\
VISARL\cite{liang2024visarl}           & 0.0095  &  0.0295  & 40.0 & 5.02 ± 0.47            \\
Zhang et al.\cite{zhang2025learning}   & 0.0054  &  0.0082  & 62.0 & 8.30 ± 0.22            \\ \rowcolor{gray!20} 
\textbf{Ours-Rnet50}                   & \textbf{0.0021}  &  \textbf{0.0067}  & \textbf{72.0} & \textbf{8.49 ± 0.25}            \\
\bottomrule
\end{tabular}
\end{center}
\label{table_comparison}
\end{table}

\begin{figure}[thpb]
      \centering
      \includegraphics[scale=0.66]{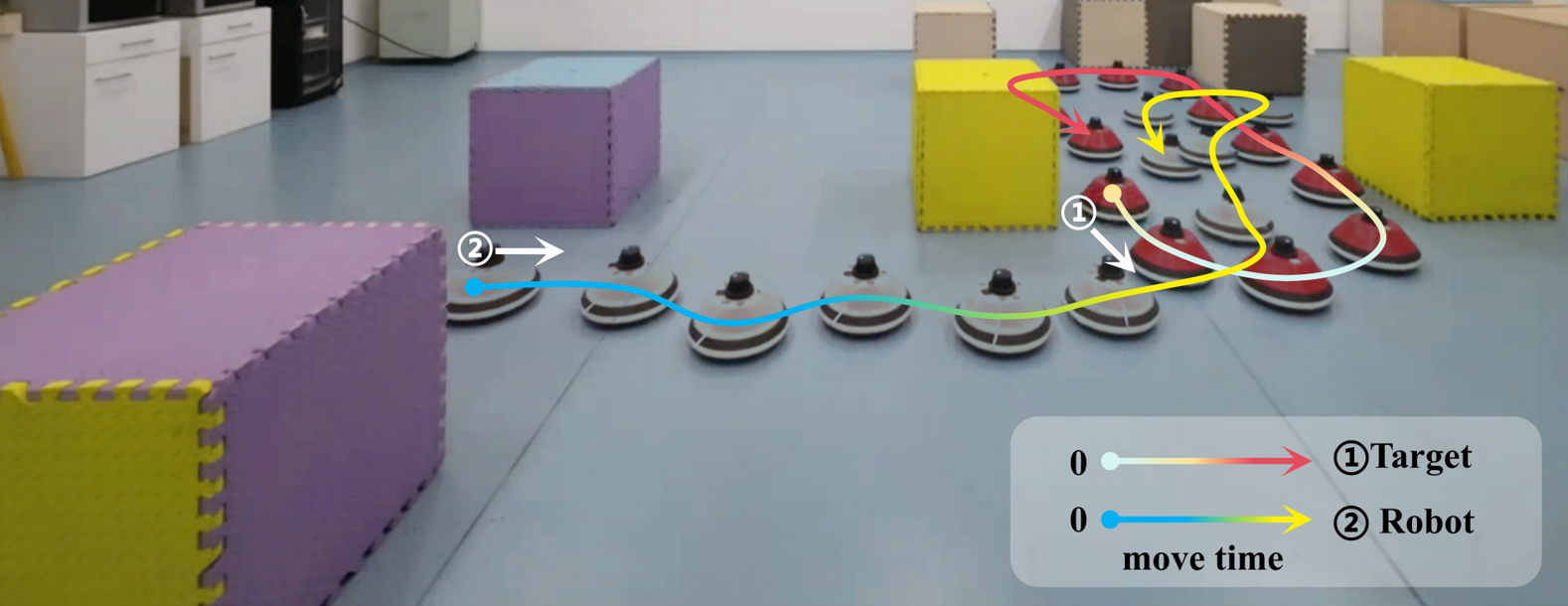}
      \caption{Robot trajectories in real-world experiment. The white is the robot and the red is the dynamic target. }
      \label{figure_real_world}
 \end{figure}
\subsection{Comparisons With Other Methods}
We compare our knowledge distillation method with open~source methods aimed at improving scene transfer and overall performance in visuomotor policies. RoboSaGA \cite{zhuang2024enhancing} and VISARL \cite{liang2024visarl} leverage visual attention maps to enhance visuomotor policies, 
while 
Zhang et al. \cite{zhang2025learning} employs contrastive learning to improve the image encoder's representation power.
Table~\ref{table_comparison} shows the results.
While both our method and \cite{zhang2025learning}'s use cross-modality feature learning, ours show better performance in our task. The reason for this is that our method jointly distills both intermediate embeddings and action outputs, enabling the visual encoder to be optimized using gradients jointly derived from the task-relevant action loss and the feature-relevant contrastive loss. In contrast, \cite{zhang2025learning} trains the cross-modality encoder solely via contrastive learning and freezes the encoder during downstream RL policy learning, which limits the visual encoder’s ability to adapt to the specific task. 
This phenomenon is consistent with the recent findings of Wang et al. \cite{wang2025feature}, which concluded that treating the image encoder as part of the policy and performing end-to-end training results in better performance.
Frozen pretrained image encoders cannot adapt their learned features to the specific visual characteristics required for the downstream action learning task, which can lead to suboptimal performance. In addition, \cite{zhang2025learning} does not consider omni-view inputs, which limits policy performance on robots with omnidirectional mobility.
Other methods\cite{zhuang2024enhancing,liang2024visarl} conduct enhancement only in monocular vision domain, without the appearance-invariant depth and omni-view information. 
Ours outperforms them in terms of AE, SR and MD, with a SR increase of 16.1\% compared with the strongest baseline. This proves the contribution of our method in improving the scene transfer and navigational performance for mobile robots.

\section{Conclusions}
In this paper, we propose an omni-view cross-modality knowledge distillation framework to enhance vision-based policies for mobile robots. By distilling intermediate features and action outputs from a teacher policy trained with omni-view depth observations, a monocular policy can inherit rich spatial representations, leading to improved transferability and navigation safety. Extensive simulation and real-world experiments demonstrate that our method outperforms pre-trained encoders and alternative enhancement approaches, achieving approximately 20\% higher success rates.





\bibliographystyle{ieeetr}
\bibliography{ref3}
\end{document}